\pdfoutput=1

\documentclass[11pt]{article}

\usepackage[preprint]{acl}

\usepackage{hyperref}
\usepackage{times}
\usepackage{latexsym}
\usepackage{tabularx}
\usepackage{multirow}
\usepackage{adjustbox}
\usepackage{fvextra}
\usepackage{bbding}

\usepackage[T1]{fontenc}

\usepackage[utf8]{inputenc}

\usepackage{microtype}

\usepackage{inconsolata}

\usepackage{graphicx}

\title{PUCP-Metrix: An Open-source and Comprehensive Toolkit for Linguistic Analysis of Spanish Texts}

\author{Javier Alonso Villegas Luis$\dagger$ \and Marco Antonio Sobrevilla Cabezudo$\dagger$$\ddagger$ \\
$\dagger$Research Group on Artificial Intelligence, Pontificia Universidad Católica del Perú \\ 
$\ddagger$Aveni \\
\texttt{\{alonso.villegas, msobrevilla\}@pucp.edu.pe}
}

\begin{document}
\maketitle
\begin{abstract}
Linguistic features remain essential for interpretability and tasks that involve style, structure, and readability, but existing Spanish tools offer limited coverage. We present PUCP-Metrix, an open-source and comprehensive toolkit for linguistic analysis of Spanish texts. PUCP-Metrix includes 182 linguistic metrics spanning lexical diversity, syntactic and semantic complexity, cohesion, psycholinguistics, and readability. It enables fine-grained, interpretable text analysis. We evaluate its usefulness on Automated Readability Assessment and Machine-Generated Text Detection, showing competitive performance compared to an existing repository and strong neural baselines. PUCP-Metrix offers a comprehensive and extensible resource for Spanish, supporting diverse NLP applications.
\end{abstract}

\section{Introduction}

Linguistic features have gained renewed importance in explainable NLP, particularly for tasks requiring interpretability, stylistic sensitivity, or attention to surface-level properties. Despite advances in end-to-end neural models, recent work shows that handcrafted or derived features remain essential in applications such as AI-generated text detection \citep{kumarage2023stylometricdetectionaigeneratedtext,ciccarelli-etal-2024-team,petukhova-etal-2024-petkaz}, educational NLP \citep{MIZUMOTO2023100050,pmlr-v273-hou25a,AtkinsonPalma2025}, and readability assessment \citep{Zeng_Tong_Yu_Xiao_Huang_2024,7339cd3d89074a95a3acb4afcc8a6bc9}. In automated essay scoring, for instance, models incorporating linguistic features offer more transparent and pedagogically meaningful evaluations \citep{pmlr-v273-hou25a}. These trends highlight the need for robust, modular repositories/toolkits that allow to extract linguistic metrics that complement deep models.

Beyond NLP applications, these toolkits also support linguistic research, offering standardized, quantifiable descriptions of texts across genres, registers, and proficiency levels \citep{Jiang2016,Kuiken+2023+83+93}. They enable empirical analyses of morphosyntactic variation, cohesion \cite{GonzalezDiosBengoetxea2021_VaxxStance}, or lexical sophistication \cite{CROSSLEY201846}, and facilitate cross-linguistic studies \cite{Ucar2024}.

Existing tools have demonstrated the value of this approach. For instance, Coh-Metrix \citep{McNamara05052010} provides extensive metrics for English across various linguistic levels. Similar resources include NILC-Metrix for Portuguese \citep{NILC_METRIX}, Coh-Metrix-Esp for Spanish \citep{quispesaravia-etal-2016-coh}, and MultiAzterTest for Spanish, English, and Basque \citep{bengoetxea2021multiaztertest}. Nevertheless, the latter two exhibit limited metric coverage and present challenges related to installing and inference efficiency.

In this work, we introduce PUCP-Metrix, a new open-source toolkit for extracting linguistic metrics from Spanish texts. It extends coverage to lexical, syntactic, discourse, psycholinguistic, and readability dimensions, providing a total of 182 linguistic metrics. In addition, we demonstrate its utility in two downstream tasks: Automated Readability Assessment and Machine-Generated Text Detection.

Our main contributions are:
\begin{itemize}
\item PUCP-Metrix, a comprehensive and extensible open-source toolkit of linguistic metrics for Spanish, featuring metrics not available in existing resources.\footnote{The code is available at \url{https://github.com/iapucp/pucp-metrix}.}
\item An empirical study evaluating its usefulness in Automated Readability Assessment and Machine-Generated Text Detection.
\end{itemize}





\section{PUCP-Metrix}

PUCP-Metrix is a modular and extensible linguistic analysis toolkit for Spanish, designed to support both research and large-scale text processing through a Python library. Its architecture emphasizes flexibility and scalability, enabling users to efficiently extract a wide range of linguistic metrics from texts.

To achieve this, we leveraged the widely adopted NLP library spaCy for core processing tasks such as tokenization, part-of-speech tagging, and dependency parsing. Building on spaCy’s modular architecture, we developed custom pipelines that include both general-purpose and category-specific metrics implemented as reusable components. This design allows external users to easily extend or modify the system, ensuring that PUCP-Metrix remains both efficient and adaptable to new linguistic analyses.

\subsection{Linguistic Categories and Metrics}

We employed an open-source Spanish implementation of \textit{Coh-Metrix} \cite{quispesaravia-etal-2016-coh} to collect initial linguistic metrics and guide our design. To develop additional metrics, we examined the implementations provided by tools such as MultiAzterTest and NILC-Metrix. Overall, we compiled  182 linguistic metrics for Spanish texts. The complete list is available at Appendix~\ref{sec:appendix_pucp_metrix_metrics}.
    
\begin{itemize}
    \item \textbf{Descriptives}: 27 indicators that capture general statistics of the text, such as \textit{number of words}, \textit{number of sentences}, \textit{number of paragraphs}, \textit{ minimum and maximum length of sentences}, \textit{average word length}.
    
    \item \textbf{Lexical Diversity}: 
    22 indicators measure the diversity of the text’s vocabulary, including the \textit{type-token ratio for various word categories (nouns, verbs, etc.)}, \textit{noun density}, \textit{verb density}, \textit{adverb density}, and \textit{adjective density}. Our implementation extends these measures with type-token ratios for additional word categories and their lemmatized forms. Another key indicators include the following:
\begin{itemize}
    \item \textit{MTLD (Measure of Textual Lexical Diversity)}: Addresses TTR's length sensitivity by calculating the average length of sequential word segments that maintain a certain TTR threshold, providing more stable measures across varying text lengths \cite{mccarthy2010}.
    \item \textit{VOCd (Vocabulary Complexity Diversity)}: Estimates vocabulary richness through curve-fitting techniques on random samples, offering insights into the probability of encountering new word types \cite{mccarthy2010}.
    \item \textit{Maas Index}: A logarithmic transformation that provides an alternative measure of lexical diversity, particularly useful for comparing texts of different lengths \cite{mass1972zusammenhang}.
\end{itemize}
    
    \item \textbf{Readability}: 7 indicators that represent how difficult to understand the text is, such as \textit{Flesch Grade Level}, \textit{Brunet Index}, \textit{Gunning Fog Index}, \textit{Honore's Statistic}, \textit{SMOG Grade}, \textit{The Szigriszt-Pazos Perspicuity Index} and \textit{Readability µ}.Among the important measures are:

    \begin{itemize}
    \item \textit{Flesch Grade Level (Fernández-Huertas adaptation)}: Adapted for Spanish texts, this measure estimates the grade level required for comprehension. 

    \item \textit{Brunet Index}: A readability measure of lexical richness, where lower values indicate greater vocabulary diversity.

    \item \textit{Gunning Fog Index}: Calculates readability by considering both sentence length and complex word percentage, estimating the education level needed for comprehension.

    \item \textit{Honore's Statistic}: Measures vocabulary richness by analyzing hapax legomena (words appearing only once).

    \item \textit{SMOG Grade}: Estimates the years of education required to understand a text by analyzing polysyllabic words (3+ syllables).
    
    \item \textit{Szigriszt-Pazos Perspicuity Index}: A Spanish-specific readability measure that evaluates text clarity, offering insights into Spanish text comprehensibility.
    
    \item \textit{Readability µ}: A statistical measure that evaluates text complexity through letter distribution patterns.
\end{itemize}
    
    \item \textbf{Syntactic Complexity}: 12 indicators, reflecting the structural intricacy of text, such as \textit{proportion of sentences with 1-7 clauses}, \textit{minimal edit distances of words, POS tags and lemmas}. Following \textit{Coh-Metrix}, our implementation extends minimal edit distance measures to POS tags and lemmatized forms, providing comprehensive syntactic variation analysis.

    \item \textbf{Psycholinguistics}:  30 indicators, reflecting psycholinguistic properties of words, specifically how they are understood by humans: \textit{concreteness}, \textit{imageability}, \textit{familiarity}, \textit{age of acquisition}, \textit{valence} and \textit{arousal}. These psycholinguistic properties were collected from the EsPal database \cite{duchon2013} and works from \citet{gonzalez2017}:

    \begin{itemize}

    \item \textit{Concreteness}: Measures the degree to which words refer to tangible, physical objects versus abstract concepts. Higher concreteness values indicate words that are easier to visualize and process cognitively.
    
    \item \textit{Imageability}: Assesses how easily words can evoke mental images. Words with higher imageability are processed more quickly and remembered more easily.
    \item \textit{Familiarity}: Evaluates how well-known words are to speakers. Familiar words are processed faster and require less cognitive effort.
    
    \item \textit{Age of Acquisition}: Measures the age at which words are typically learned. Earlier acquired words are processed more automatically and efficiently.
    \item \textit{Valence}: Assesses the emotional positivity or negativity of words. Valence influences emotional processing and memory formation.
    
    \item \textit{Arousal}: Measures the emotional intensity or activation level of words. Arousal affects attention and memory consolidation.
\end{itemize}

    \item \textbf{Word Information}:  24 indicators with more detailed word-level statistics, such as: \textit{number of nouns}, \textit{number of verbs}, \textit{number of adverbs}, \textit{number of adjectives} and \textit{number of content words}.

    \item \textbf{Referential Cohesion}: 12 indicators that serve to measure the interconnections within a text: \textit{noun overlap}, \textit{argument overlap}, \textit{stem overlap}, \textit{content word overlap} and \textit{anaphor overlap}.

    \item \textbf{Textual Simplicity}: 4 indicators, measuring the simplicity of the text using the ratio of short or large sentences, such as: \textit{proportion of short sentences}, \textit{proportion of medium sentences}, \textit{proportion of long sentences}, \textit{proportion of very long sentences}.

    \item \textbf{Semantic Cohesion}: 8 indicators, assessing the degree of semantic relatedness between different parts of the text, such as: \textit{Latent Semantic Analysis (LSA) overlap of adjacent sentences}, \textit{LSA overlap of all sentences}, \textit{LSA overlap of adjacent paragraphs}.

    \item \textbf{Word Frequency}: 16 indicators, various measurements involving the Zipf's frequency\footnote{Zipf's frequency is estimate by using the wordfreq tool. It is available at \url{https://github.com/rspeer/wordfreq/}.} for different kinds of words, such as \textit{rare nouns count}, \textit{rare verbs count}, \textit{rare adverbs count}, \textit{rare content words count}\footnote{Rare words were defined in a similar way as \citet{bengoetxea2021multiaztertest}.} and \textit{mean word frequency}.
    
    \item \textbf{Syntactic Pattern Density}: 14 indicators, reflecting the density of various syntactic elements, such as: \textit{noun phrase density}, \textit{verb phrase density}, \textit{negative expressions density}, \textit{coordinating conjunctions density} and \textit{subordinating conjunction density}.

    \item \textbf{Connectives}: 6 indicators, measuring the use of words or phrases that establish logical, temporal, or other relationships between different parts of the text, such as: \textit{casual connectives incidence}, \textit{logical connectives incidence}, \textit{adversative connectives incidence}, \textit{temporal connectives incidence}, \textit{additive connectives incidence}, \textit{all connectives incidence}.
\end{itemize}




\subsection{Comparison with Existing Tools}





Table~\ref{tab:tools_comparison} compares Coh-Metrix-Esp, MultiAzterTest, and PUCP-Metrix across 3 dimensions: ease of installation, inference time (i.e., the time required to extract linguistic features), and metric coverage. PUCP-Metrix improves upon existing tools by combining straightforward installation, fast inference, and a broad set of linguistic metrics. In contrast, MultiAzterTest, while providing many metrics, is challenging to install and depends on external tools such as syntactic parsers, resulting in slower feature extraction and limited scalability. PUCP-Metrix, by offering 182 metrics—significantly more than Coh-Metrix-Esp (48) and MultiAzterTest (141)—provides a practical and comprehensive solution for research in computational linguistics.

\begin{table}[h!]
\scalebox{0.85}{
\begin{tabular}{lp{1cm}p{1.5cm}p{1.5cm}}
\hline
                   & Easy to install               & Inference Time & Metric Coverage \\ \hline
Coh-Metrix-Esp     & \XSolidBrush   & Fast           & 48              \\
MultiAzterTest     & \XSolidBrush   & Slow           & 141             \\
PUCP-Metrix (ours) & \CheckmarkBold & Fast           & 182             \\ \hline
\end{tabular}
}
\caption{Comparison of PUCP-Metrix with existing tools for Spanish.}

\label{tab:tools_comparison}
\end{table}

Table~\ref{tab:metrics_per_tool} shows the number of linguistic metrics implemented in Coh-Metrix-Esp, MultiAzterTest and PUCP-Metrix (ours). PUCP-Metrix provides a broader coverage of linguistic metrics compared to Coh-Metrix-Esp and MultiAzterTest,  across 13 categories. Notably, PUCP-Metrix includes metrics in categories that are entirely missing or underrepresented in the other tools, such as semantic cohesion, textual simplicity, and psycholinguistics. This way, PUCP-Metrix can capture higher-level discourse, cognitive readability, and psycholinguistic properties.

Furthermore, PUCP-Metrix distributes its metrics more evenly across lexical, syntactic, semantic, and psycholinguistic dimensions. This comprehensive and balanced coverage allows for a more detailed and nuanced characterization of texts, making PUCP-Metrix better suited for in-depth linguistic analysis and a wide range of NLP applications.

\begin{table*}[ht!]
\centering
\scalebox{0.9}{
\begin{tabular}{l|ccc}
\hline
Category                  & Coh-Metrix-Esp & MultiAzterTest & PUCP-Metrix (ours) \\ \hline
Descriptive               & 11            & 22             & \textbf{27}         \\
Referential Cohesion      & 12            & 10             & 12         \\
Lexical Diversity         & 2             & 20             & \textbf{22}         \\
Readability               & 1             & 1              & \textbf{7}          \\
Connectives               & 6             & 12             & 6          \\
Syntactic Complexity      & 2             & 19             & 12         \\
Pattern Density           & 3             & 0              & \textbf{14}         \\
Semantic Cohesion         & 0             & 0              & \textbf{8}          \\
Word Information          & 11            & 32             & 24         \\
Word Frequency            & 0             & 15             & \textbf{16}         \\
Textual Simplicity        & 0             & 0              & \textbf{4}          \\
Psycholinguistics         & 0             & 0              & \textbf{30}         \\
Word Semantic Information & 0             & 4              & 0          \\
Semantic Overlap          & 0             & 6              & 0          \\ \hline
Total                     & 48            & 141            & 182        \\ \hline
\end{tabular}
}
\caption{Number of linguistic metrics per category for each tool.}
\label{tab:metrics_per_tool}
\end{table*}

\section{Applications}

In order to verify the usefulness of PUCP-Metrix, we use it for two tasks where linguistic metrics have proven to be helpful in past work. In particular, we select Automated Readability Assessment (ARA) and Machine-Generated Text Detection.

\subsection{Automated Readability Assessment (ARA)}

We adopt an approach similar to that of \citet{vasquez-rodriguez-etal-2022-benchmark}, who introduced a benchmark for ARA on Spanish texts. Their work unified both public and non-public corpora annotated with language proficiency levels and proposed two- and three-class classification schemas.

In contrast, our study comprises only four publicly available datasets —CAES, Coh-Metrix-Esp, Kwiziq, and HablaCultura— to ensure reproducibility and open accessibility. We adopt the same label mappings described in the paper, adapting all texts to two readability classification schemas: 2-label (simple, complex) and 3-label (basic, intermediate, advanced). The dataset's descriptions and the labeling strategy can be found in Appendix~\ref{sec:ara_datasets}.

Overall, the dataset contains 32,167 instances, distributed across the four sources as follows: 31,149 from CAES, 100 from Coh-Metrix-Esp, 206 from Kwiziq, and 713 from HablaCultura.

We experiment with two readability classification schemas mentioned before. All experiments are performed at the document level\footnote{We use the same texts that come from the available resources}. The corpus is divided into 80\% training, 10\% validation, and 10\% test sets, stratified by label. We evaluate models using Precision, Recall, and F1-score.

\subsection{Machine-Generated Text Detection}

We adopt the AuTexTification 2023 shared task dataset \cite{autex2023}, which comprises over 160,000 texts in English and Spanish across five domains: tweets, reviews, news, legal, and how-to articles generated by both human and large language models.

For our experiments, we focus on the Machine-generated Text Detection task, which consists of identifying if a text has been created by a human or a machine. The task includes 26,996 human-generated instances and 25,195 machine-generated instances, totaling 52,191 instances. More details about the dataset can be found in Appendix~\ref{sec:ara_datasets}.

\subsection{Models}

For both tasks, we trained several machine learning models—Logistic Regression (LR), XGBoost (XGB), Support Vector Machines (SVM), and Random Forests (RF)—using the metrics extracted with both MultiAzterTest and PUCP-Metrix.

Following the AuTexTification shared task setup and consistent with the ARA formulation, we use a RoBERTa-based model \cite{fandinoetal_2022} (RoBERTa-BNE)\footnote{Model available at \url{https://huggingface.co/PlanTL-GOB-ES/roberta-base-bne}}
 as our baseline. We fine-tune this model on the official training splits and evaluate it on the corresponding test splits to ensure comparability. Similarly, we fine-tune RoBERTa-BNE on both the 2-label and 3-label ARA classification schemas.

\section{Results and Discussion}

\subsection{Automated Readability Assessment}

Table~\ref{tab:results_complex} reports the results for the 2-label ARA task. PUCP-Metrix slightly outperforms MultiAzter, achieving an overall F1 of 97.46 with XGBoost compared to 97.24. However, this difference is not significant. XGBoost consistently yields the highest scores, followed by Random Forest, while Logistic Regression and SVM lag behind. RoBERTa-BNE achieves the best overall F1 (98.30), indicating that deep contextual models capture subtle semantic patterns beyond what feature-based metrics provide.

\begin{table}[]
\centering
\resizebox{\columnwidth}{!}{
\begin{tabular}{llllllll}
\hline
\multicolumn{1}{c|}{\multirow{2}{*}{Model}} &
  \multicolumn{3}{c|}{Simple} &
  \multicolumn{3}{c|}{Complex} &
  \multicolumn{1}{c}{\multirow{2}{*}{F1}} \\
\multicolumn{1}{c|}{} &
  \multicolumn{1}{c}{P} &
  \multicolumn{1}{c}{R} &
  \multicolumn{1}{c|}{F1} &
  \multicolumn{1}{c}{P} &
  \multicolumn{1}{c}{R} &
  \multicolumn{1}{c|}{F1} &
  \multicolumn{1}{c}{} \\ \hline
\multicolumn{8}{c}{Multiazter}                                                                                             \\ \hline
\multicolumn{1}{l|}{LR}  & 96.42 & 97.27 & \multicolumn{1}{l|}{96.85} & 91.75 & 89.37 & \multicolumn{1}{l|}{90.54} & 93.70 \\
\multicolumn{1}{l|}{XGB} & 98.05 & 99.20 & \multicolumn{1}{l|}{98.62} & 97.57 & 94.19 & \multicolumn{1}{l|}{95.85} & 97.24 \\
\multicolumn{1}{l|}{SVM} & 96.51 & 97.32 & \multicolumn{1}{l|}{96.91} & 91.89 & 89.62 & \multicolumn{1}{l|}{90.74} & 93.82 \\
\multicolumn{1}{l|}{RF}  & 97.25 & 99.29 & \multicolumn{1}{l|}{98.26} & 97.76 & 91.72 & \multicolumn{1}{l|}{94.64} & 96.45 \\ \hline
\multicolumn{8}{c}{PUCP-Metrix}                                                                                            \\ \hline
\multicolumn{1}{l|}{LR}  & 96.68 & 97.65 & \multicolumn{1}{l|}{97.16} & 92.87 & 90.11 & \multicolumn{1}{l|}{91.47} & 94.31 \\
\multicolumn{1}{l|}{XGB} & 98.38 & 99.08 & \multicolumn{1}{l|}{98.73} & 97.22 & 95.18 & \multicolumn{1}{l|}{96.19} & 97.46 \\
\multicolumn{1}{l|}{SVM} & 96.60 & 97.69 & \multicolumn{1}{l|}{97.14} & 92.97 & 89.86 & \multicolumn{1}{l|}{91.39} & 94.27 \\
\multicolumn{1}{l|}{RF}  & 97.45 & 99.20 & \multicolumn{1}{l|}{98.32} & 97.52 & 92.34 & \multicolumn{1}{l|}{94.86} & 96.59 \\ 
\hline
\multicolumn{1}{l|}{RoBERTa-BNE}  & 99.04 & 99.24 & \multicolumn{1}{l|}{99.14} & 97.76 & 97.16 & \multicolumn{1}{l|}{97.46} & 98.30 \\ 
\hline
\end{tabular}}
\caption{Results on 2-label ARA/Complexity Classification task}
\label{tab:results_complex}
\end{table}

Table~\ref{tab:results_complex_multi} shows the 3-label ARA results. PUCP-Metrix again slightly surpasses MultiAzter (F1 96.72 vs. 96.56), with XGBoost as the top-performing classifier. RoBERTa-BNE remains the strongest model, achieving an overall F1 of 98.13 and near-perfect performance on Basic and Intermediate texts.

\begin{table*}[]
\centering
\scalebox{0.9}{
\begin{tabular}{lllllllllll}
\hline
\multicolumn{1}{c|}{\multirow{2}{*}{Model}} &
  \multicolumn{3}{c|}{Basic} &
  \multicolumn{3}{c|}{Intermediate} &
  \multicolumn{3}{c|}{Advanced} &
  \multicolumn{1}{c}{\multirow{2}{*}{F1}} \\
\multicolumn{1}{c|}{} &
  \multicolumn{1}{c}{P} &
  \multicolumn{1}{c}{R} &
  \multicolumn{1}{c|}{F1} &
  \multicolumn{1}{c}{P} &
  \multicolumn{1}{c}{R} &
  \multicolumn{1}{c|}{F1} &
  \multicolumn{1}{c}{P} &
  \multicolumn{1}{c}{R} &
  \multicolumn{1}{c|}{F1} &
  \multicolumn{1}{c}{} \\ \hline
\multicolumn{11}{c}{Multiazter}                                                                                                                                         \\ \hline
\multicolumn{1}{l|}{LR}  & 91.43 & 92.56 & \multicolumn{1}{l|}{91.99} & 85.66 & 86.30 & \multicolumn{1}{l|}{85.97} & 83.00 & 74.20 & \multicolumn{1}{l|}{78.36} & 85.44 \\
\multicolumn{1}{l|}{XGB} & 97.62 & 98.59 & \multicolumn{1}{l|}{98.10} & 96.43 & 96.59 & \multicolumn{1}{l|}{96.51} & 98.48 & 91.87 & \multicolumn{1}{l|}{95.06} & 96.56 \\
\multicolumn{1}{l|}{SVM} & 90.54 & 93.08 & \multicolumn{1}{l|}{91.79} & 85.29 & 85.71 & \multicolumn{1}{l|}{85.50} & 84.72 & 68.55 & \multicolumn{1}{l|}{75.78} & 84.36 \\
\multicolumn{1}{l|}{RF}  & 96.32 & 98.07 & \multicolumn{1}{l|}{97.18} & 94.38 & 94.93 & \multicolumn{1}{l|}{94.66} & 98.37 & 85.16 & \multicolumn{1}{l|}{91.29} & 94.38 \\ \hline
\multicolumn{11}{c}{PUCP-Metrix}                                                                                                                                        \\ \hline
\multicolumn{1}{l|}{LR}  & 92.25 & 92.85 & \multicolumn{1}{l|}{92.55} & 86.35 & 86.71 & \multicolumn{1}{l|}{86.53} & 82.02 & 77.39 & \multicolumn{1}{l|}{79.64} & 86.24 \\
\multicolumn{1}{l|}{XGB} & 97.68 & 98.59 & \multicolumn{1}{l|}{98.13} & 97.16 & 96.59 & \multicolumn{1}{l|}{96.88} & 96.72 & 93.64 & \multicolumn{1}{l|}{95.15} & 96.72 \\
\multicolumn{1}{l|}{SVM} & 91.10 & 93.55 & \multicolumn{1}{l|}{92.31} & 86.06 & 85.63 & \multicolumn{1}{l|}{85.85} & 82.72 & 71.02 & \multicolumn{1}{l|}{76.43} & 84.86 \\
\multicolumn{1}{l|}{RF}  & 95.55 & 98.18 & \multicolumn{1}{l|}{96.85} & 95.11 & 93.77 & \multicolumn{1}{l|}{94.44} & 97.63 & 87.28 & \multicolumn{1}{l|}{92.16} & 94.48 \\ \hline
\multicolumn{1}{l|}{RoBERTa-BNE}  & 99.30 & 99.24 & \multicolumn{1}{l|}{99.27} & 98.83 & 98.42 & \multicolumn{1}{l|}{98.63} & 95.50 & 97.53 & \multicolumn{1}{l|}{96.50} & 98.13 \\ \hline
\end{tabular}}
\caption{Results on 3-label ARA/Complexity Classification task}
\label{tab:results_complex_multi}
\end{table*}  

\subsection{Machine-Generated Text Detection}

Table~\ref{tab:results_autex} reports the performance of machine learning models using PUCP-Metrix and MultiAzter metrics, alongside a RoBERTa-BNE model fine-tuned on AuTexTification and the shared task’s best-reported results.

PUCP-Metrix consistently outperforms MultiAzter. For human texts, F1 increases from 42–51 (MultiAzter) to 60–66, and for machine texts from 70–73 to 71–76, showing its ability to capture linguistic and structural cues. Tree-based models, especially XGBoost and Random Forest, benefit most, achieving the highest overall F1.

Compared to RoBERTa-BNE, PUCP-Metrix provides more balanced class performance. While RoBERTa-BNE attains high precision on human texts (86.28), low recall (46.87) limits its F1, suggesting that contextual embeddings may miss the diversity of human writing. PUCP-Metrix also slightly surpasses the shared task’s best-reported F1 (70.77), indicating that integrating linguistic features with neural models could further improve results.

Finally, we analyze which linguistic metrics contribute most to classification. Overall, detecting machine-generated text depends primarily on features related to frequency, readability, and cohesion, whereas ARA tasks are driven by descriptive, syntactic, and simplicity-related features. Full details of the feature analysis are provided in Appendix~\ref{sec:feat_analysis}.

\begin{table}[]
\centering
\resizebox{\columnwidth}{!}{
\begin{tabular}{cccccccc}
\hline
\multicolumn{1}{l|}{\multirow{2}{*}{Model}} & \multicolumn{3}{c|}{Human} & \multicolumn{3}{c|}{Machine} & \multirow{2}{*}{F1} \\
\multicolumn{1}{c|}{}    & P     & R     & \multicolumn{1}{c|}{F1}    & P     & R     & \multicolumn{1}{c|}{F1}    &       \\ \hline
\multicolumn{8}{c}{Multiazter}                                                                                             \\ \hline
\multicolumn{1}{l|}{LR}  & 70.52 & 30.28 & \multicolumn{1}{c|}{42.37} & 61.84 & 89.93 & \multicolumn{1}{c|}{73.29} & 57.83 \\
\multicolumn{1}{l|}{XGB} & 68.10 & 39.73 & \multicolumn{1}{c|}{50.18} & 63.98 & 85.19 & \multicolumn{1}{c|}{73.08} & 61.63 \\
\multicolumn{1}{l|}{SVM} & 70.43 & 30.74 & \multicolumn{1}{c|}{42.80} & 61.95 & 89.73 & \multicolumn{1}{c|}{73.30} & 58.05 \\
\multicolumn{1}{l|}{RF}  & 62.08 & 43.98 & \multicolumn{1}{c|}{51.49} & 63.82 & 78.62 & \multicolumn{1}{c|}{70.45} & 60.97 \\ \hline
\multicolumn{8}{c}{PUCP-Metrix}                                                                                            \\ \hline
\multicolumn{1}{l|}{LR}  & 71.09 & 55.93 & \multicolumn{1}{c|}{62.61} & 70.02 & 81.90 & \multicolumn{1}{c|}{75.49} & 69.05 \\
\multicolumn{1}{l|}{XGB} & 71.34 & 61.36 & \multicolumn{1}{c|}{\textbf{65.97}} & 72.33 & 80.38 & \multicolumn{1}{c|}{76.14} & \textbf{71.06} \\
\multicolumn{1}{l|}{SVM} & 71.04 & 56.05 & \multicolumn{1}{c|}{62.66} & 70.06 & 81.82 & \multicolumn{1}{c|}{75.48} & 69.07 \\
\multicolumn{1}{l|}{RF}  & 63.57 & 58.24 & \multicolumn{1}{c|}{60.79} & 68.85 & 73.44 & \multicolumn{1}{c|}{71.07} & 65.93 \\ \hline
\multicolumn{1}{l|}{RoBERTa-BNE}  & 86.28 & 46.87 & \multicolumn{1}{c|}{60.74} & 68.99 & 94.07 & \multicolumn{1}{c|}{79.60} & 70.17 \\
\multicolumn{1}{l|}{RoBERTa-Autex*}  & - & - & \multicolumn{1}{c|}{-} & - & - & \multicolumn{1}{c|}{-} & 68.52 \\
\multicolumn{1}{l|}{Best model*}  & - & - & \multicolumn{1}{c|}{-} & - & - & \multicolumn{1}{c|}{-} & 70.77 \\ \hline
\end{tabular}}
\caption{Results on AuTexTification. *The authors of the shared task only provide F1 in the report.}
\label{tab:results_autex}
\end{table}






\section{Tool Usage}

PUCP-Metrix can be installed via \texttt{pip}:

\begin{verbatim}
pip install iapucp-metrix
\end{verbatim}

To use the library, we need to import the \texttt{Analyzer} class and call \texttt{compute\_metrics} to compute all metrics. The function supports multiprocessing through spaCy, allowing us to specify the number of workers and the batch size.

\begin{Verbatim}[fontsize=\small]
from iapucp_metrix.analyzer import Analyzer

analyzer = Analyzer()

texts = ["Este es mi ejemplo."]

metrics_list = analyzer.compute_metrics(
    texts, 
    workers=4,
    batch_size=2
)

for i, metrics in enumerate(metrics_list):
    print(Readability (Fernández-Huertas):)
    print(f"{metrics['RDFHGL']:.2f}")
\end{Verbatim}

The output of the code described above is:
\begin{Verbatim}[fontsize=\small]
Readability (Fernández-Huertas):
201.86
\end{Verbatim}

In addition, PUCP-Metrix supports computing metrics grouped by linguistic categories (via \texttt{compute\_grouped\_metrics}), enabling users to analyze model behavior across dimensions such as lexical, syntactic, and semantic features.

\section{Conclusion and Future Work}
    

PUCP-Metrix provides a linguistically rich set of 182 metrics for Spanish, offering broader coverage and a larger metric set than previous resources. Empirical evaluations demonstrate its effectiveness in ARA and Machine-generated text detection tasks. Models trained on these metrics match or outperform baseline neural models, underscoring their ability to capture nuanced linguistic information.

Future work includes expanding the metric set to incorporate more discourse and pragmatic metrics, adapting PUCP-Metrix to other Spanish varieties, and integrating these metrics into pre-trained language models or NLP pipelines. Benchmarking on larger and more diverse tasks/datasets will further validate its robustness and support the development of specialized metric sets.

\section*{Limitations}

The current evaluation has several limitations. Although PUCP-Metrix has been tested on multiple datasets, the experiments primarily focus on learner essays, children’s texts, and selected AuTexTification domains, leaving its performance on other genres and domains uncertain. Additionally, PUCP-Metrix depends heavily on spaCy-based linguistic processing and external lexicons (e.g., psycholinguistic norms), so parsing errors and coverage gaps in these resources can directly affect the reliability of the computed metrics.


\bibliography{custom}

\appendix

\section{List of metrics in PUCP-Metrix}
\label{sec:appendix_pucp_metrix_metrics}

\begin{table*}[]
\begin{adjustbox}{width=\textwidth}
\begin{tabular}{cl|cl}
\hline
Category                                       & \multicolumn{1}{c|}{Metric Description}                                  & Category                                            & \multicolumn{1}{c}{Metric Description}                                  \\ \hline
\multirow{27}{*}{Descriptive Indices}          & DESPC: Paragraph count                                                   & \multirow{12}{*}{Syntactic Complexity Indices}      & SYNNP: Mean number of modifiers per noun phrase                         \\
                                               & DESPCi: Paragraph count incidence per 1000 words                         &                                                     & SYNLE: Mean number of words before main verb                            \\
                                               & DESSC: Sentence count                                                    &                                                     & SYNMEDwrd: Minimal edit distance of words between adjacent sentences    \\
                                               & DESSCi: Sentence count incidence per 1000 words                          &                                                     & SYNMEDlem: Minimal edit distance of lemmas between adjacent sentences   \\
                                               & DESWC: Word count (alphanumeric words)                                   &                                                     & SYNMEDpos: Minimal edit distance of POS tags between adjacent sentences \\
                                               & DESWCU: Unique word count                                                &                                                     & SYNCLS1: Ratio of sentences with 1 clause                               \\
                                               & DESWCUi: Unique word count incidence per 1000 words                      &                                                     & SYNCLS2: Ratio of sentences with 2 clauses                              \\
                                               & DESPL: Average paragraph length (sentences per paragraph)                &                                                     & SYNCLS3: Ratio of sentences with 3 clauses                              \\
                                               & DESPLd: Standard deviation of paragraph length                           &                                                     & SYNCLS4: Ratio of sentences with 4 clauses                              \\
                                               & DESSL: Average sentence length (words per sentence)                      &                                                     & SYNCLS5: Ratio of sentences with 5 clauses                              \\
                                               & DESSLd: Standard deviation of sentence length                            &                                                     & SYNCLS6: Ratio of sentences with 6 clauses                              \\
                                               & DESSNSL: Average sentence length excluding stopwords                     &                                                     & SYNCLS7: Ratio of sentences with 7 clauses                              \\ \cline{3-4} 
                                               & DESSNSLd: Standard deviation of sentence length excluding stopwords      & \multirow{14}{*}{Syntactic Pattern Density Indices} & DRNP: Noun phrase density per 1000 words                                \\
                                               & DESSLmax: Maximum sentence length                                        &                                                     & DRNPc: Noun phrase count                                                \\
                                               & DESSLmin: Minimum sentence length                                        &                                                     & DRVP: Verb phrase density per 1000 words                                \\
                                               & DESWLsy: Average syllables per word                                      &                                                     & DRVPc: Verb phrase count                                                \\
                                               & DESWLsyd: Standard deviation of syllables per word                       &                                                     & DRNEG: Negation expression density per 1000 words                       \\
                                               & DESCWLsy: Average syllables per content word                             &                                                     & DRNEGc: Negation expression count                                       \\
                                               & DESCWLsyd: Standard deviation of syllables per content word              &                                                     & DRGER: Gerund form density per 1000 words                               \\
                                               & DESCWLlt: Average letters per content word                               &                                                     & DRGERc: Gerund count                                                    \\
                                               & DESCWLltd: Standard deviation of letters per content word                &                                                     & DRINF: Infinitive form density per 1000 words                           \\
                                               & DESWLlt: Average letters per word                                        &                                                     & DRINFc: Infinitive count                                                \\
                                               & DESWLltd: Standard deviation of letters per word                         &                                                     & DRCCONJ: Coordinating conjunction density per 1000 words                \\
                                               & DESWNSLlt: Average letters per word (excluding stopwords)                &                                                     & DRCCONJc: Coordinating conjunction count                                \\
                                               & DESWNSLltd: Standard deviation of letters per word (excluding stopwords) &                                                     & DRSCONJ: Subordinating conjunction density per 1000 words               \\
                                               & DESLLlt: Average letters per lemma                                       &                                                     & DRSCONJc: Subordinating conjunction count                               \\ \cline{3-4} 
                                               & DESLLltd: Standard deviation of letters per lemma                        & \multirow{6}{*}{Connective Indices}                 & CNCAll: All connectives incidence per 1000 words                        \\ \cline{1-2}
\multirow{7}{*}{Readability Indices}           & RDFHGL: Fernández-Huertas Grade Level                                    &                                                     & CNCCaus: Causal connectives incidence per 1000 words                    \\
                                               & RDSPP: Szigriszt-Pazos Perspicuity                                       &                                                     & CNCLogic: Logical connectives incidence per 1000 words                  \\
                                               & RDMU: Readability µ index                                                &                                                     & CNCADC: Adversative connectives incidence per 1000 words                \\
                                               & RDSMOG: SMOG index                                                       &                                                     & CNCTemp: Temporal connectives incidence per 1000 words                  \\
                                               & RDFOG: Gunning Fog index                                                 &                                                     & CNCAdd: Additive connectives incidence per 1000 words                   \\ \cline{3-4} 
                                               & RDHS: Honoré Statistic                                                   & \multirow{24}{*}{Word Information Indices}          & WRDCONT: Content word incidence per 1000 words                          \\
                                               & RDBR: Brunet index                                                       &                                                     & WRDCONTc: Content word count                                            \\ \cline{1-2}
\multirow{12}{*}{Referential Cohesion Indices} & CRFNO1: Noun overlap between adjacent sentences                          &                                                     & WRDNOUN: Noun incidence per 1000 words                                  \\
                                               & CRFAO1: Argument overlap between adjacent sentences                      &                                                     & WRDNOUNc: Noun count                                                    \\
                                               & CRFSO1: Stem overlap between adjacent sentences                          &                                                     & WRDVERB: Verb incidence per 1000 words                                  \\
                                               & CRFCWO1: Content word overlap between adjacent sentences (mean)          &                                                     & WRDVERBc: Verb count                                                    \\
                                               & CRFCWO1d: Content word overlap between adjacent sentences (std dev)      &                                                     & WRDADJ: Adjective incidence per 1000 words                              \\
                                               & CRFANP1: Anaphore overlap between adjacent sentences                     &                                                     & WRDADJc: Adjective count                                                \\
                                               & CRFNOa: Noun overlap between all sentences                               &                                                     & WRDADV: Adverb incidence per 1000 words                                 \\
                                               & CRFAOa: Argument overlap between all sentences                           &                                                     & WRDADVc: Adverb count                                                   \\
                                               & CRFSOa: Stem overlap between all sentences                               &                                                     & WRDPRO: Personal pronoun incidence per 1000 words                       \\
                                               & CRFCWOa: Content word overlap between all sentences (mean)               &                                                     & WRDPROc: Personal pronoun count                                         \\
                                               & CRFCWOad: Content word overlap between all sentences (std dev)           &                                                     & WRDPRP1s: First person singular pronoun incidence per 1000 words        \\
                                               & CRFANPa: Anaphore overlap between all sentences                          &                                                     & WRDPRP1sc: First person singular pronoun count                          \\ \cline{1-2}
\multirow{22}{*}{Lexical Diversity Indices}    & LDTTRa: Type-token ratio for all words                                   &                                                     & WRDPRP1p: First person plural pronoun incidence per 1000 words          \\
                                               & LDTTRcw: Type-token ratio for content words                              &                                                     & WRDPRP1pc: First person plural pronoun count                            \\
                                               & LDTTRno: Type-token ratio for nouns                                      &                                                     & WRDPRP2s: Second person singular pronoun incidence per 1000 words       \\
                                               & LDTTRvb: Type-token ratio for verbs                                      &                                                     & WRDPRP2sc: Second person singular pronoun count                         \\
                                               & LDTTRadv: Type-token ratio for adverbs                                   &                                                     & WRDPRP2p: Second person plural pronoun incidence per 1000 words         \\
                                               & LDTTRadj: Type-token ratio for adjectives                                &                                                     & WRDPRP2pc: Second person plural pronoun count                           \\
                                               & LDTTRLa: Type-token ratio for all lemmas                                 &                                                     & WRDPRP3s: Third person singular pronoun incidence per 1000 words        \\
                                               & LDTTRLno: Type-token ratio for noun lemmas                               &                                                     & WRDPRP3sc: Third person singular pronoun count                          \\
                                               & LDTTRLvb: Type-token ratio for verb lemmas                               &                                                     & WRDPRP3p: Third person plural pronoun incidence per 1000 words          \\
                                               & LDTTRLadv: Type-token ratio for adverb lemmas                            &                                                     & WRDPRP3pc: Third person plural pronoun count                            \\ \cline{3-4} 
                                               & LDTTRLadj: Type-token ratio for adjective lemmas                         & \multirow{30}{*}{Psycholinguistic Indices}          & PSYC: Overall concreteness ratio                                        \\
                                               & LDTTRLpron: Type-token ratio for pronouns                                &                                                     & PSYC0: Very low concreteness ratio (1-2.5)                              \\
                                               & LDTTRLrpron: Type-token ratio for relative pronouns                      &                                                     & PSYC1: Low concreteness ratio (2.5-4)                                   \\
                                               & LDTTRLipron: Type-token ratio for indefinite pronouns                    &                                                     & PSYC2: Medium concreteness ratio (4-5.5)                                \\
                                               & LDTTRLifn: Type-token ratio for functional words                         &                                                     & PSYC3: High concreteness ratio (5.5-7)                                  \\
                                               & LDMLTD: Measure of Textual Lexical Diversity (MTLD)                      &                                                     & PSYIM: Overall imageability ratio                                       \\
                                               & LDVOCd: Vocabulary Complexity Diversity (VoCD)                           &                                                     & PSYIM0: Very low imageability ratio (1-2.5)                             \\
                                               & LDMaas: Maas index                                                       &                                                     & PSYIM1: Low imageability ratio (2.5-4)                                  \\
                                               & LDDno: Noun density                                                      &                                                     & PSYIM2: Medium imageability ratio (4-5.5)                               \\
                                               & LDDvb: Verb density                                                      &                                                     & PSYIM3: High imageability ratio (5.5-7)                                 \\
                                               & LDDadv: Adverb density                                                   &                                                     & PSYFM: Overall familiarity ratio                                        \\
                                               & LDDadj: Adjective density                                                &                                                     & PSYFM0: Very low familiarity ratio (1-2.5)                              \\ \cline{1-2}
\multirow{16}{*}{Word Frequency Indices}       & WFRCno: Rare noun count                                                  &                                                     & PSYFM1: Low familiarity ratio (2.5-4)                                   \\
                                               & WFRCnoi: Rare noun incidence per 1000 words                              &                                                     & PSYFM2: Medium familiarity ratio (4-5.5)                                \\
                                               & WFRCvb: Rare verb count                                                  &                                                     & PSYFM3: High familiarity ratio (5.5-7)                                  \\
                                               & WFRCvbi: Rare verb incidence per 1000 words                              &                                                     & PSYAoA: Overall age of acquisition ratio                                \\
                                               & WFRCadj: Rare adjective count                                            &                                                     & PSYAoA0: Very early acquisition ratio (1-2.5)                           \\
                                               & WFRCadji: Rare adjective incidence per 1000 words                        &                                                     & PSYAoA1: Early acquisition ratio (2.5-4)                                \\
                                               & WFRCadv: Rare adverb count                                               &                                                     & PSYAoA2: Medium acquisition ratio (4-5.5)                               \\
                                               & WFRCadvi: Rare adverb incidence per 1000 words                           &                                                     & PSYAoA3: Late acquisition ratio (5.5-7)                                 \\
                                               & WFRCcw: Rare content word count                                          &                                                     & PSYARO: Overall arousal ratio                                           \\
                                               & WFRCcwi: Rare content word incidence per 1000 words                      &                                                     & PSYARO0: Very low arousal ratio (1-3)                                   \\
                                               & WFRCcwd: Distinct rare content word count                                &                                                     & PSYARO1: Low arousal ratio (3-5)                                        \\
                                               & WFRCcwdi: Distinct rare content word incidence per 1000 words            &                                                     & PSYARO2: Medium arousal ratio (5-7)                                     \\
                                               & WFMcw: Mean frequency of content words                                   &                                                     & PSYARO3: High arousal ratio (7-9)                                       \\
                                               & WFMw: Mean frequency of all words                                        &                                                     & PSYVAL: Overall valence ratio                                           \\
                                               & WFMrw: Mean frequency of rarest words per sentence                       &                                                     & PSYVAL0: Very negative valence ratio (1-4)                              \\
                                               & WFMrcw: Mean frequency of rarest content words per sentence              &                                                     & PSYVAL1: Negative valence ratio (3-5)                                   \\ \cline{1-2}
\multirow{8}{*}{Semantic Cohesion Indices}     & SECLOSadj: LSA overlap between adjacent sentences (mean)                 &                                                     & PSYVAL2: Positive valence ratio (5-7)                                   \\
                                               & SECLOSadjd: LSA overlap between adjacent sentences (std dev)             &                                                     & PSYVAL3: Very positive valence ratio (7-9)                              \\ \cline{3-4} 
                                               & SECLOSall: LSA overlap between all sentences (mean)                      & \multirow{6}{*}{Textual Simplicity Indices}         &                                                                         \\
                                               & SECLOSalld: LSA overlap between all sentences (std dev)                  &                                                     & TSSRsh: Ratio of short sentences (\textless 11 words)                   \\
                                               & SECLOPadj: LSA overlap between adjacent paragraphs (mean)                &                                                     & TSSRmd: Ratio of medium sentences (11-12 words)                         \\
                                               & SECLOPadjd: LSA overlap between adjacent paragraphs (std dev)            &                                                     & TSSRlg: Ratio of long sentences (13-14 words)                           \\
                                               & SECLOSgiv: LSA overlap between given and new sentences (mean)            &                                                     & TSSRxl: Ratio of very long sentences ($\geq$ 15 words)                       \\
                                               & SECLOSgivd: LSA overlap between given and new sentences (std dev)        &                                                     &                                                                         \\ \hline
\end{tabular}
\end{adjustbox}
\label{tab:pucp_metrix_metrics}
\caption{List of linguistic metrics implemented in PUCP-Metrix}
\end{table*}

\section{Datasets for Automated Readability Assessment and Machine-generated Text Detection}
\label{sec:ara_datasets}

\subsection{Automated Readability Assessment}

\begin{itemize}
    \item CAES (\textit{Corpus de Aprendices del Español})\footnote{Available at \url{https://galvan.usc.es/caes/}} \cite{parodi2015corpus}. This corpus consists of essays written by learners of Spanish as a foreign language. Each document is annotated with a CEFR level (A1–C2). Following \citet{vasquez-rodriguez-etal-2022-benchmark}, we map A1–B1 to "simple" and B2–C2 to "complex" for the 2-label schema, and A1-A2 to "basic", B1-B2 to "intermediate" and C1-C2 to "advanced" for the 3-label schema.
    \item Coh-Metrix-Esp \cite{quispesaravia-etal-2016-coh}. This dataset is a collection of short Spanish stories that  includes children’s tales and texts intended for adults. It provides explicit simple and complex labels, directly aligned to our 2-label schema and to "basic" vs "advanced" in the 3-label schema.
    \item Kwiziq\footnote{The platform is available at \url{https://www.kwiziq.com/}}. Kwiziq is an online language-learning platform that offers graded Spanish readings labeled with CEFR levels. We use the available data proposed by \citet{vasquez-rodriguez-etal-2022-benchmark} and map the CEFR annotations to our 2- and 3-label classification schemes using the same criteria.
    \item HablaCultura. This dataset comprises educational readings sourced from the HablaCultura platform\footnote{Available at \url{https://hablacultura.com/}}, where each text is labeled by instructors with CEFR levels. We use the same level mappings used by \citet{vasquez-rodriguez-etal-2022-benchmark}.
\end{itemize}

\subsection{Machine-generated Text Detection}

Human-generated texts in AuTexTification were sourced from publicly available datasets, including MultiEURLEX \cite{chalkidis-etal-2021-multieurlex} (legal), XLSUM/MLSUM \cite{hasan-etal-2021-xl,scialom-etal-2020-mlsum} (news), COAR/COAH \cite{10.1007/978-3-319-07983-7_28} (reviews), XLM-Tweets \cite{barbieri-etal-2022-xlm} and TSD \cite{LeisTSD} (tweets), and WikiLingua \cite{ladhak-etal-2020-wikilingua} (how-to articles). Machine-generated texts were produced using six large language models: three from the BLOOM family (BLOOM-1B7\footnote{Available at \url{https://huggingface.co/bigscience/bloom-1b7}.}, BLOOM-3B\footnote{Available at \url{https://huggingface.co/bigscience/bloom-3b}.}, BLOOM-7B1\footnote{Available at \url{https://huggingface.co/bigscience/bloom-7b1}.}) and three from the GPT-3 family (babbage, curie, text-davinci-003).

\section{Feature Analysis}
\label{sec:feat_analysis}

We applied Anova over our dataset using all the metrics. We set a p-value of 0.05 and remove the features that do not make contribution for our analysis.

\begin{figure}[h!]
  \centering
  \includegraphics[width=\columnwidth]{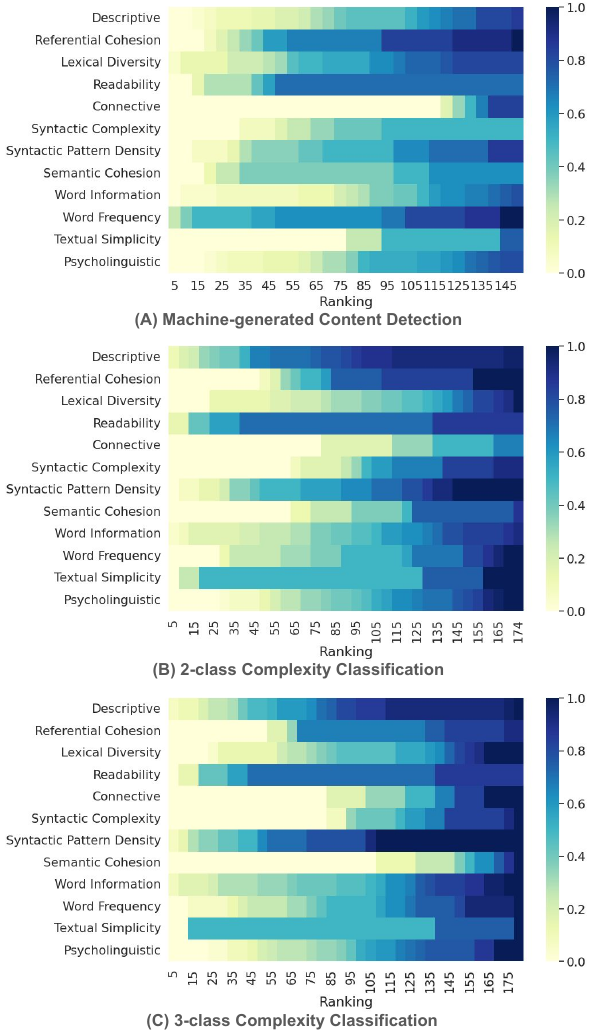}
  \caption{Category coverage along the ranking for PUCP-Metrix}
  \label{fig:coverage_category}
\end{figure}

Figure~\ref{fig:coverage_category} shows a heatmap representing the coverage of linguistic categories along the ranking, i.e., the distribution of linguistic features as more signals are included. Overall, the contribution of linguistic features varies across tasks. For machine-generated content detection, top-ranked signals are dominated by word frequency, readability, and semantic cohesion metrics. In contrast, descriptive and connective metrics play a more limited role and appear only at later ranks.

For ARA tasks, the importance shifts toward descriptive features, syntactic pattern density, readability, syntactic complexity, and textual simplicity metrics. Conversely, semantic cohesion and connective metrics are comparatively less important.

\end{document}